**Title:** Improving the quality of dental crown using a Transformer-based method

Golriz Hosseinimanesh[1], Farnoosh Ghadiri [2], Ammar Alsheghri[1], Ying Zhang [1], Julia Keren [3], Farida Cheriet [1], Francois Guibault [1]
1- Department of Computer and Software Engineering, École Polytechnique de Montréal, Montréal, QC, Canada
2- Centre d'intelligence artificielle appliquée (JACOBB), Montréal, QC, Canada
3- Intellident Dentaire Inc., Montréal, QC, Canada

**Abstract**
Designing a synthetic crown is a time-consuming, inconsistent, and labor-intensive process. In this work, we present a fully automatic method that not only learns human design dental crowns, but also improves the consistency, functionality, and esthetic of the crowns. Following success in point cloud completion using the transformer-based network, we tackle the problem of the crown generation as a point-cloud completion around a prepared tooth. To this end, we use a geometry-aware transformer to generate dental crowns. Our main contribution is to add a margin line information to the network, as the accuracy of generating a precise margin line directly,determines whether the designed crown and prepared tooth are closely matched to allowappropriateadhesion.Using our ground truth crown, we can extract the margin line as a spline and sample the spline into 1000 points. We feed the obtained margin line along with two neighbor teeth of the prepared tooth and three closest teeth in the opposing jaw. We also add the margin line points to our ground truth crown to increase the resolution at the margin line. Our experimental results show an improvement in the quality of the designed crown when considering the actual context composed of the prepared tooth along with the margin line compared with a crown generated in an empty space as was done by other studies in the literature.Our code is available at :
"https://github.com/Golriz-code/shellGeneration/tree/main/Shell%20Generation"

## 1. Introduction

If a tooth is missing, decayed, or fractured, it may need a dental crown. Professional dental practitioners need a significant amount of effort, skill, and time to design a personalized crown for each patient with current digital technologies for dentistry, such as computer-aided design software. Besides, there are functional and geometric constraints that put limits on possible solutions and desired precision. In a nutshell, the designed crown should not overlap existing teeth, and it must respect a space occlusion relationship with the opposing teeth to provide adequate chewing function. Each dental crown consists of a crown shell, crown bottom, and margin line. Here, our goal is to generate the crown shell using a fully automated algorithm, given a prep and its surrounding (context).

*1.1 Related work:*

Few studies have been conducted on automating dental crown design. The author [1] down-scaled 3D dental scans into 2D depth images and then created depth images of dental crowns in a way similar to pix2pix paper [2]. In this way, they may make use of improvements in image generation to achieve good performance. But information is lost during their conversion from 3D to 2D. Ping et al [3] presented SA-IFN, a voxel-based method for completing 3D tooth scans that integrate the functions of implicit representations with the self-attention mechanism. Voxel-based methods, on the other hand, need several unnecessary occupancy predictions, which reduces the output resolution. Zhu et al. [4] applied a new voxel-based reconstruction after completing the tooth point clouds using a transformer-based model. However, as opposed to reconstructing a thin surface with a boundary, their technique creates a watertight mesh.

3D deep learning has significantly advanced in recent years. Due to the frequent partial nature of 3D data coming from inaccurate scans, point cloud completion is essential for 3D computer vision. Predicting an object's complete shape from a partial observation is a challenging task. This problem has received increasing attention in the computer vision community since the pioneering works of PointNet and FoldingNet [5]. As mentioned in Fei et al. [6], models can be categorized into point-based, convolution-based, graph-based, folding-based, GANbased, and transformer-based methods. Among these architectures, transformers [7] suggest a unique attention mechanism that uses many levels of self-attention and cross-multi-head attention to interchange data between input and output. Although the attention mechanism was initially intended for NLP tasks, it has recently been demonstrated to be beneficial for a variety of computer vision problems. Each layer of the transformer then consists of an attention mechanism, which allows for interaction between inputs at different positions, followed by a position-wise fully connected network, which is applied to all positions independently.

*1.2 New work to be presented:*

There is currently no intelligent technology that can automatically assist technicians in generating dental crowns for prepared teeth, also referred to as 'preps'. In this paper, we propose a 3D shape completion transformer-based network to automatically reconstruct a patient-specific dental crown shape in which the dental crown fits onto the prepared tooth perfectly at the margin line. Our contributions are summarized as follows: 1) We propose to add the margin line information to our ground truth crown to increase the resolution at the margin line; 2) Our network is able to generate teeth in any position. In Contrast with [8] which only trained their network for the specific tooth, our network is able to generate teeth in any position.

## 2. Methods

Our approach uses an end-to-end deep learning architecture to generate the dental crown from given partial arches. We train our network on all incisors, canines, molars, and premolars crowns.

## *2.1 Data preprocessing:*

We first create a context from a given 3D scan. Context refers to the two neighbor teeth of a prepared tooth and the three teeth in the opposing jaw. Figure 1 shows the input data and created context. To create a context for a specifically prepared tooth, a 3D scan is semantically segmented into 14 classes (number of teeth in each jaw) using MeshSegNet method [9] trained on our dataset. Then two teeth in the prep's neighbor and the three closest teeth in the opposing jaw were selected. We also add gingiva triangles close to the selected teeth. In the next step, we extract the margin line from a die (i.e., a scan of a prepared tooth). Based on the available crown cases for which we have the crown geometries provided by dental technicians using the commercial software of Exocad, we were able to extract the exact ground truth margin lines that were generated by the technicians. As opposed to other work published in the literature [8], we used the prepared tooth geometry and the ground truth margin line to enhance the fitting of the generated crown on the prepared tooth. By having more localized point density at the margin line we also teach the network to place a higher weight at the margin line. In production, we are using our pre-trained MeshSegNet to segment margin lines from prepared teeth [9] [10]. Figure 2 shows the ground truth shell with and without margin line.

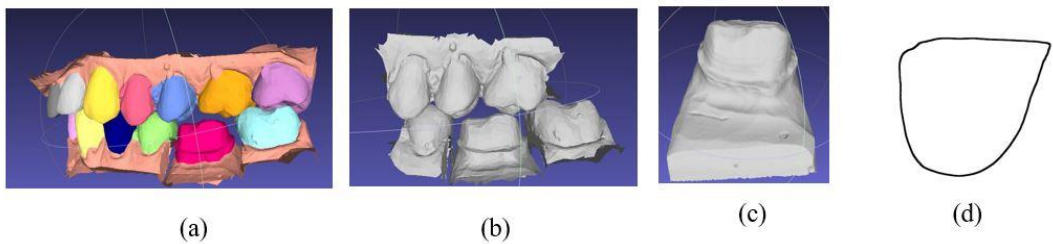

Figure 1: (a) Dental arch with a prep, (b) Generated context from segmentation, (c) die (prepped tooth) (d) Extracted margin line.

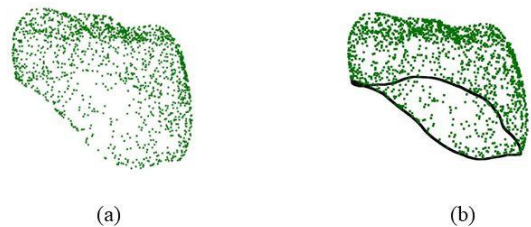

Figure 2: (a) Ground truth shell without margin line, (b) Ground truth shell with margin line.

## 2.2 Network Architecture:

We use a transformer encoder-decoder architecture proposed by [11] to predict the point proxies for the dental crown. Figure 3 illustrates our proposed framework for a point-based dental crown generation. The transformer-based model by [11] tackles the point cloud completion problem as a set-to-set translation problem. The model is characterized by five main components: an encoder-decoder architecture to convert point cloud completion as a set-to-set translation problem, point proxies to represent the original point clouds as a set of feature vectors, geometry-aware transformer block to facilitate transformers to better leverage the inductive bias about 3D geometric structures, query generator to dynamically generate queries for the decoder, and multi-scale point cloud generation process to generate final prediction in a course to fine manner.

One of the key challenges of applying transformers is that the self-attention mechanism in transformers lacks some inductive biases that explicitly model the structure of the point cloud. To enable transformers to better leverage the inductive bias about the 3D geometric structure of point clouds, [11] proposed a geometry-aware block that models the geometric relations which can be a plug-and-play module is designed to incorporate the attention blocks with any transformer architecture. The multi-scale point could generation is done by a small modification in the original FoldingNet [5] which takes predicted proxies, and centers from the transformer, and tries to map the 3D grid onto a proper tooth by mimicking the deformation of a 3D point cloud. We also used three consecutive folding operations, instead of two in the original FoldingNet paper, to help produce detailed structures.

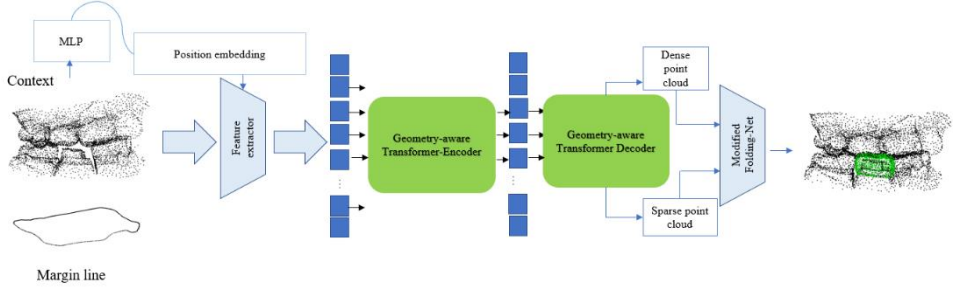

Figure 3: The pipeline of our proposed network.

We employed the mean chamfer distance as the evaluation metric, which can measure the distance between the prediction point cloud and ground truth at a set level. Fitting the crown with the neighboring teeth can be controlled by this metric. For each prediction, the chamfer distance between the prediction point set *p* and the ground truth point set *g* is calculated.

$$d_{CD}(P,G) = \frac{1}{|p|}\sum_{p \epsilon P} \min_{g \epsilon G} \min||p-g||_2^2 + \frac{1}{|g|}\sum_{g \epsilon G} \min_{g \epsilon G} \min||g-p||_2^2 \quad Eq.\ (1)$$

Where P, G $\in R^{N*3}$ are point sets with N points, $||\ldots||_2$ means the 2-Norm of a vector.

## 3. Experimental results

Our experimental dataset contains 90 training, 10 validation, and 25 test cases. The dataset included all type of teeth in the jaw, namely molars, canines, and incisors. We uniformly sampled 1568 points from the crown and 10240 points for the context and 1000 points for the margin line. Our training procedure is like the one used in [11]. The input point clouds were normalized by subtracting the mean of the points and dividing the outcome by the standard deviation. Regular augmentation methods such as rotation, translation, and scaling were applied to meshes. Figure 4 shows our results qualitatively. To evaluate our generation, we used the Chamfer distance (CD) between the prediction point set and the ground-truth point set. The mean CD of the different experiments is shown in Table 1.

Table 1: The mean CD of the different experiments.

|  | Experiment | CD-L1 | CD-L2 |
|---|---|---|---|
| All positions | Baseline + Margin line | $56.39 * 10^{-3}$ | $8.04 * 10^{-3}$ |

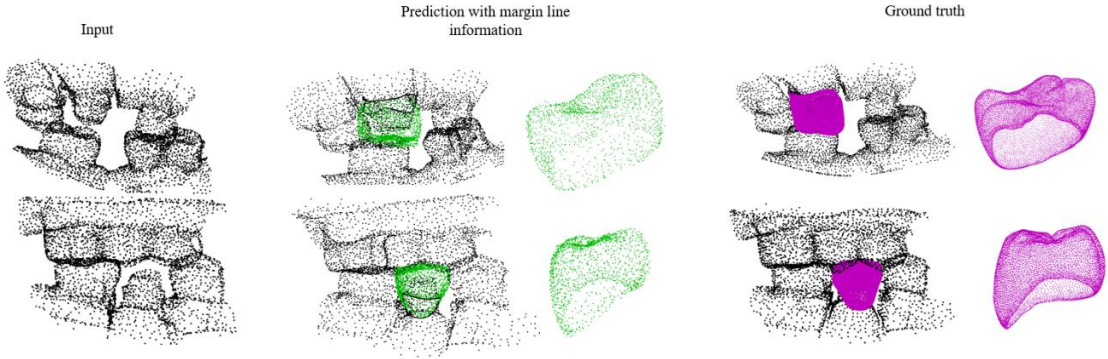

Figure 4: Prediction with margin line information.

In this stage, our goal is to evaluate the margin line that we can extract from the reconstructed surface of the predicted shell. We use a simple reconstruction method such as ball pivoting algorithm [12] (it can be substitute with other surface reconstruction methods such as Poisson surface reconstruction) to reconstruct the predicted shell. Then the margin line is extracted from the reconstructed surface . Figure 5 shows the predicted shell , reconstructed shell and extracted marginline. In this experiment, we first predict the shell by the transformer-based network without considering margin line information. Second, we predict the shell by considering margin line information in the shell and context input. Finally, we evaluate the extracted margin line from reconstructed shell of the two experiments.

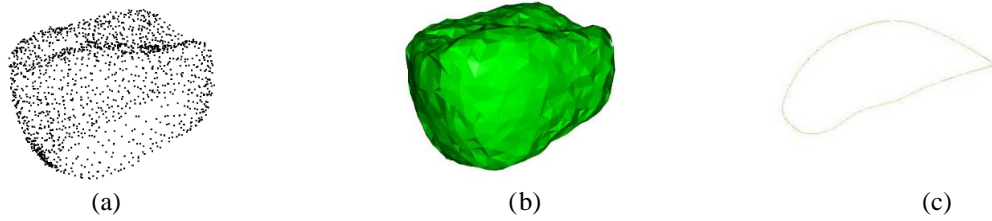

(a) (b) (c)

Figure 5: This figure shows the point predicted shell (a) , and the reconstructed surface of the shell (b), the extracted margin line (c).

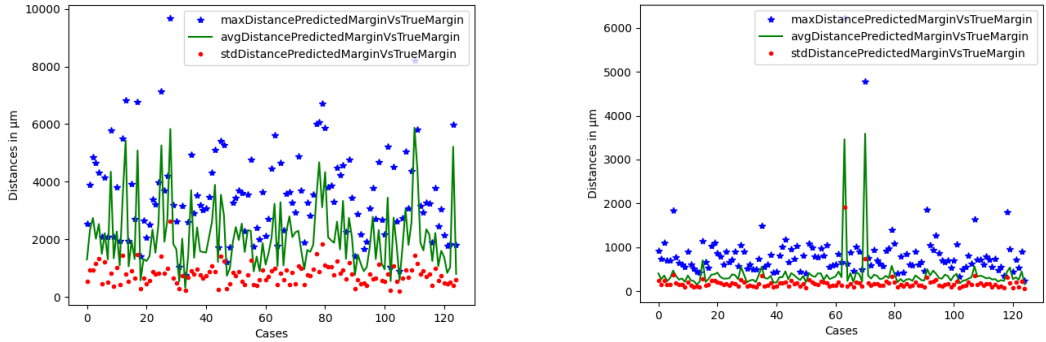

Figure 6: The diagram shows the (max, avg, std) distance of extracted margin line from the predicted shell and ground truth margin line. The left one is without considering margin line information in the input shell and context, the right one is by considering the margin line information.

The diagrams in figure 6 show the comparison of the distance between extracted margin line from the predicted shell and the ground truth margin line. We notice that when we consider the margin line information, there is much improvement in the predicted margin line in comparison to the ground truth margin line. There are some quantitative results in table 2.

| Metrics | Without margin line information | With margin line information |
|---|---|---|
| max distance | 3501.1311 | 847.5652 |
| min distance | 695.0081 | 60.1825 |
| Average distance | 2105.3221 | 372.2782 |
| Std distance | 783.3138 | 184.4864 |

Table 2: The average distance of the predicted margin line Vs ground truth margin line. (Distance are in $\mu m$.)

We evaluate the margin line extracted from the predicted shell with and without margin line information.

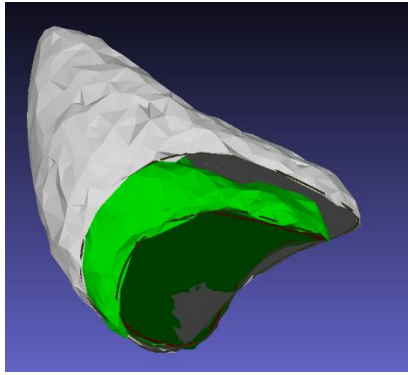

Figure 7: This figure shows the predicted crowns, the green one with margin line information and the grey one without margin line information. The black line of points is the prediction for the margin line and the red line point is the ground truth margin line.

Figure 7 shows one qualitative result in which one shell is predicted once by considering margin line information in the shell and context input, the green one, and the second without this information, the grey one. We can see the improvement in the scale of the predicted shell and the extracted margin line is so close to the ground truth margin line.

### 4. Implementation Details.

We implement our networks on PyTorch. All modules in our completion network are trained with an ADAM optimizer, and the initial learning rate is set to 0.0005 with a decay rate of 0.9 every 20 epochs. We set our batch size to 16 and the number of workers to 8. We assign a radius in the range [0.6, 0.7, 0.8, 0.9, 1] for ball pivoting surface reconstruction.

### 5. Conclusions

Here, in contrast to recent works on dental restorations which focused only on restoring plausible missing teeth, we generated a crown for a realistic setting with the prepared tooth. We have shown that by adding margin line information obtained from the prep, our model can generate better-fit crowns. In the future, we are planning to generate the shell in a mesh format by using the graph convolutional network.

### 6. Acknowledgments

This work was funded by KerenOr, Intellident Dentaire Inc., iMD Research, the Natural Science and Engineering Research Council of Canada, Institut de valorisation des données (IVADO), and MEDTEQ. We thank Compute Canada for providing the computational resources used in this work. The authors acknowledge the help and support from JACOBB and Object Research Systems Inc. This work has not been submitted for publication or presentation elsewhere.